\documentclass[runningheads,a4paper]{llncs}

\usepackage{amssymb}
\usepackage{graphicx}
\usepackage{amsmath}
\usepackage{bm}

\usepackage{url}
\urldef{\mailsa}\path|{m181021, tkurita}@hiroshima-u.ac.jp|

\begin{document}

\mainmatter  


\title{Mixup of Feature Maps in a Hidden Layer for Training of Convolutional Neural Network}

\titlerunning{Mixup of Feature Maps in a Hidden Layer}

%
%
\author{Hideki Oki \and Takio Kurita}
%
\authorrunning{Hideki Oki and Takio Kurita}

\institute{Department of Information Engineering\\
Graduate School of Engineering\\
Hiroshima university\\
\mailsa\\
}

%
%

\toctitle{Mixup of Feature Maps in a Hidden Layer}
\tocauthor{Hideki Oki and Takio Kurita}
\maketitle

\begin{abstract}
The deep Convolutional Neural Network (CNN) became very popular as a fundamental technique for image classification and objects recognition.
To improve the recognition accuracy for the more complex tasks, deeper networks have being introduced. 
However, the recognition accuracy of the trained deep CNN drastically decreases for the samples which are obtained from the outside regions of the training samples. 
To improve the generalization ability for such samples, Krizhevsky et al. proposed to generate additional samples through transformations from the existing samples and to make the training samples richer. 
This method is known as data augmentation. 
Hongyi Zhang et al. introduced data augmentation method called mixup which achieves state-of-the-art performance in various datasets.
Mixup generates new samples by mixing two different training samples.
Mixing of the two images is implemented with simple image morphing.
In this paper, we propose to apply mixup to the feature maps in a hidden layer.
To implement the mixup in the hidden layer we use the Siamese network or the triplet network architecture to mix feature maps.
From the experimental comparison, it is observed that the mixup of the feature maps obtained from the first convolution layer is more effective than the original image mixup. 
\end{abstract}

\section{Introduction}

After the deep Convolutional Neural Network (CNN) proposed by Krizhevsky et al. \cite{Krizhevsky2012} won the ILSVRC 2012 with higher score than the conventional methods, it became very popular for image classification and object recognition. 
Usually the parameters of the deep CNN are estimated by minimizing the empirical loss defined on the training samples.
Since the number of parameters in the deep CNN is very large, the use of the regularization techniques is usually necessary. 
Also, the prediction accuracy of the trained deep CNN drastically decreases when the samples obtained from 
the outside regions of the training samples \cite{Szegedy2014}.

To improve the generalization ability for such samples, Krizhevsky et al.\cite{Krizhevsky2012} proposed to generate the additional samples through transformations from the existing samples
and to make the training samples richer. 
This method is known as data augmentation.
For example small shifts in location, small rotations or shears, changes in intensity, changes in stroke thickness, changes in size etc. are used to generate the additional samples for handwritten character recognition because the labels should be invariant to such perturbations.

It is also possible to incorporate the invariance directly into a classification
function. 
Simard et al.\cite{Simard1998} proposed a modification of the error back-propagation algorithm to train the transformation invariant classification function. 
The algorithm is called tangent propagation in which the invariance
are learned by gradient descent.

We can virtually generate the perturbation by introducing additive noises in the hidden layers in the neural networks \cite{Kurita1994,Sabri2017}. Inayoshi et al. proposed to combine the neural network classifier with auto-encoder \cite{Inayoshi2005}. The sum of the squared reconstruction errors of the auto-encoder is minimized in addition to the supervised objective function for classification while injecting noise in the hidden layer of the auto-encoder. By introducing the reconstruction error and the injected noise, we can virtually generate the perturbation along the principal directions of the variations in the training samples.

Recently Hongyi Zhang et al. \cite{mixup} introduced an data augmentation method called mixup in which new samples are generated by mixing pairs of the training samples.
Mixing of the two images is implemented with simple image morphing.
This simple mixup achieves the state-of-the art performance in various datasets.

In this paper, we propose to apply mixup to the feature maps in a hidden layer instead of the input images. 
To apply mixup to the feature maps in a hidden layer, we have to extract the features in the hidden layer for a pair of the training samples.
In the proposed method, we use the Siamese Network \cite{Bromely1993,Chopra2005,Hadsell2006} or the triplet network \cite{triplet} architecture to extract the feature maps in the hidden layer.


The effectiveness of the mixup of the feature maps in a hidden layer is confirmed by the experiments on the image classification using CIFAR-10 dataset.

\section{Related Works}

\subsection{Deep Convolutional Neural Network}


The deep convolutional neural network (CNN) is effective for image classification tasks.
The computation within the convolution layers is regarded as a filtering process of the input image as
\begin{align}
f_{p,q}^{(l)}=h(\sum^{convy-1}_{s=0}\sum^{convx-1}_{t=0}w^{(l)}_{s,t}f^{(l-1)}_{p+s, q+t}+b^{(l)}) \; ,
\end{align}
where $w^{(l)}_{s,t}$ is the weight of the neuron indexed as $(s,t)$ in the $l$-th convolution layer and $b^{(l)}$ is the bias of the $l$-th convolution layer. 
The size of the convolution filter is given $convx \times convy$. The activation function of each neuron is denoted as $h$. 

Usually, pooling layers are added after the convolution layers. The pooling layer performs downsampling for reducing computational costs and enhancing against micro position changes. 

Fully-connected layers like multi layer Perceptron is connected to the convolution layers which is used to construct the classifier.

\subsection{Mixup}

\begin{figure}[ht]
\begin{center}
\includegraphics[scale=0.7]{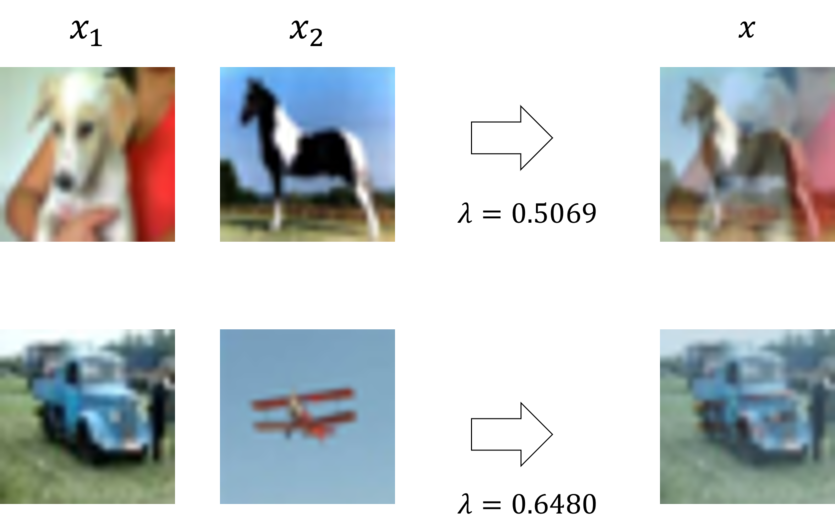}
\end{center}
\caption{mixup applied for training samples}
\end{figure}

Mixup was introduced by Hongyi Zhang et al. \cite{mixup} as an data augmentation method.
The samples are generated by mixing two different training samples by simple weighted average.
It is reported that this simple method achieves the state-of-the art performance in various datasets \cite{mixup}.
And, the similar method was introduced by Yuji Tokozume et al. \cite{blic,blic2}.
They conducted a detailed analysis of this method.

Let $\bm{x}_1$ and $\bm{x}_2$ be the images randomly extracted from the training samples and $\bm{t}_1$ and $\bm{t}_2$ are the corresponding teacher vectors.
Then the new image  $\bm{x}$ and the corresponding teacher vector $\bm{t}$ is generated by 
\begin{align} \label{mixup}
\bm{x} &= \lambda \bm{x}_1 + (1-\lambda) \bm{x}_2 \\
\bm{t} &= \lambda \bm{t}_1 + (1-\lambda) \bm{t}_2
\end{align}
where $\lambda$ is random number generated from the beta distribution $\beta(\alpha, \alpha)$ $(0 \leq \lambda \leq 1)$, and $\alpha$ is a hyper parameter ($\alpha>0$). 
The random number $\lambda$ is generated for each training pair.


It is notice that the new image $\bm{x}$ is generated as a linear interpolation of the two images $\bm{x}_1$ and $\bm{x}_2$.
It is expected that the linear interpolation gives incentives for smooth network operation and can successfully interpolate pair of the samples. 
The teacher vectors are similarly mixed by the linear interpolation.

Therefore, intermediate classes can be represented by a mixed teacher vector when the classes of two samples are different. 

We can consider mixup for three or more samples by a weighted interpolation with random numbers generated from the Dirichlet distribution.
New sample from three or more samples can be generated as
\begin{align}
\bm{x} &= u_1 \bm{x}_1 + u_2 \bm{x}_2 + u_3 \bm{x}_3 + \cdots \\
\bm{t} &= u_1 \bm{t}_1 + u_2 \bm{t}_2 + u_3 \bm{t}_3 + \cdots
\end{align}
where $u_1, u_2, u_3, \ldots $ are random numbers generated from the Dirichlet distribution $Dir(\alpha, \alpha, \alpha, \cdots)$ ($u_1 + u_2 + u_3 + \cdots = 1$), and $\alpha$ is a hyper parameter ($\alpha>0$).
It is reported that mixing of three or more samples works well as the mixing of two samples but the computational cost increases \cite{mixup}.

In this paper, we apply the weighted linear interpolation for the feature maps in a hidden layer instead of the original input images. 
To do this, we have to extract the feature maps in the hidden layer.
This is possible by using Siamese Network.

\subsection{Siamese Network and Triplet Network}

\begin{figure}[ht]
\begin{center}
\includegraphics{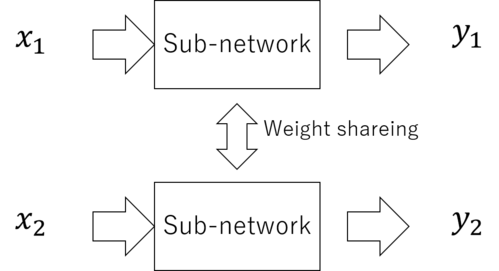}
\caption{Siamese Network}
\label{fig:siamese}
\end{center}
\end{figure}

The Siamese Network \cite{Bromely1993,Chopra2005,Hadsell2006} consists of two identical sub-networks joined at their outputs as shown in Figure \ref{fig:siamese}.
The two sub-networks extract feature vectors from two different samples.
The objective function of the optimization for training the parameters of the networks is defined by using these extracted feature vectors.
This network architecture is usually used for metric learning and the contrastive loss is often used for error function.
The parameters of the Siamese Network are trained to distinguish between similar and dissimilar pairs of the training samples. 
Usually a contrastive loss over the metric defined on the trained representation is used as the objective function for the optimization.

E. Hoffer et al. \cite{triplet} extended the Siamese Network to the network with three identical sub-networks. The network is called as a Triplet Network.
By using three sub-networks, we can treat three input images simultaneously and extract better feature vector than the Siamese Network.

In this paper, we use these architectures to extract feature maps in a hidden layer.

\section{Mixup of Feature Maps in a Hidden Layer}

\begin{figure}[ht]
\begin{center}
\includegraphics[scale=0.55]{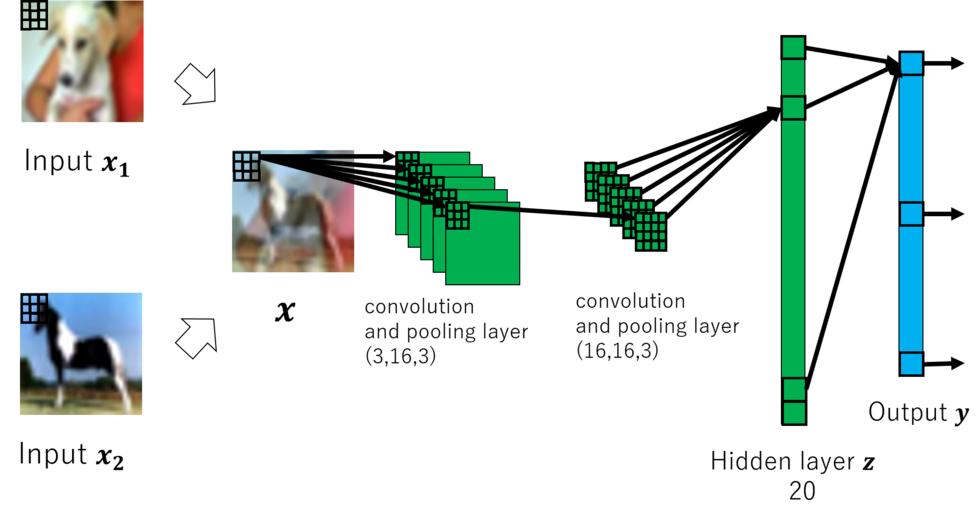}
\caption{conventional mixup}
\label{fig:mixup}
\end{center}
\begin{center}
\includegraphics[scale=0.55]{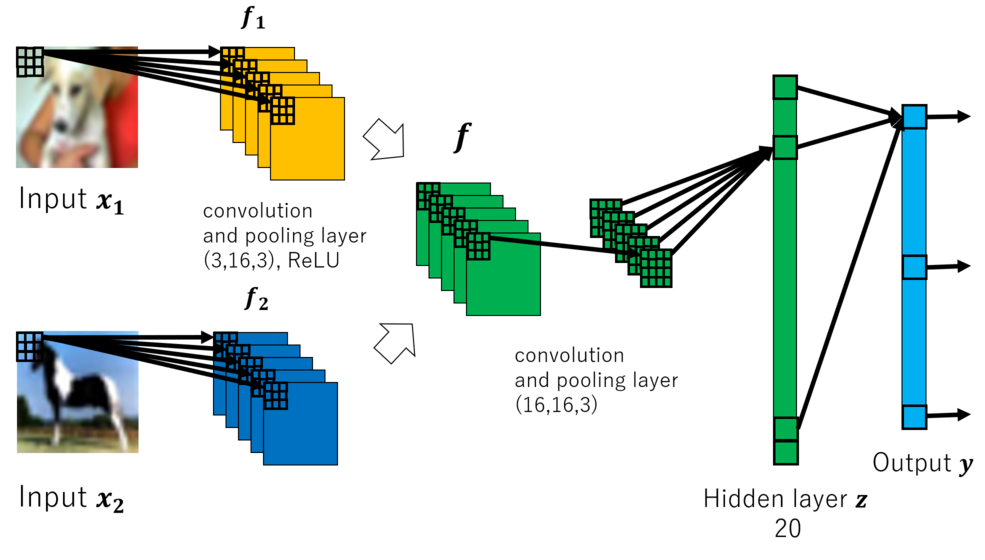}
\caption{Apply mixup to 1st convolution layer}
\label{fig:mixup-1conv}
\end{center}
\end{figure}
The mixup proposed by H. Zhang et al. generates intermediate images by mixing the pairs of the original training images as shown in Figure \ref{fig:mixup}. 
However, the original image often contains irrelevant information that is not necessary for image classification such as brightness changes or difference in subtle color. 
Mixing such irrelevant information may cause some bad influence for classification.
Since the convolution layers are trained to extract class-specific feature maps, the important information for classification are extracted in the feature map. 
This means that irrelevant information such as the brightness of the image and the difference in subtle color are reduced in the feature maps in the hidden layers.
Therefore, we propose to apply a mixup to the feature map extracted from the hidden layers. 

In order to extract feature maps of a pair of images in the training phase, we use Siamese network architecture which has two sub-networks with sharing weights as shown in Figure \ref{fig:mixup-1conv}.
For the pair of the input images $\bm{x}_1$ and $\bm{x}_2$, each of the sub-networks computes the feature maps $\bm{f}_1$ and $\bm{f}_2$ respectively.
Then these feature maps are mixed with the mixing weight $\lambda$ as
\begin{align}
\bm{f} = \lambda \bm{f}_1 + (1-\lambda) \bm{f}_2 \; .
\end{align}
Then the mixed feature map $\bm{f}$ is used as the input of the third CNN as shown in Figure \ref{fig:mixup-1conv}.
The mixing parameter $\lambda$ is generated as a random number from the $\beta$ distribution defined by
\begin{align}
\beta(x;\alpha, \alpha)=\frac{x^{\alpha-1}(1-x)^{\alpha-1}}{\displaystyle \int_{0}^{1}x^{\alpha-1}(1-x)^{\alpha-1}dx} \; .
\end{align}
The corresponding teacher vectors $\bm{t}_1$ and $\bm{t}_2$ are also mixed with the same mixing weight $\lambda$ as
\begin{align}
\bm{t} = \lambda \bm{t}_1 + (1-\lambda) \bm{t}_2 \; .
\end{align}

The weights of the sub-networks and the third CNN of the proposed network architecture are trained by minimizing the cross-entropy loss defined by using the mixed feature maps $\bm{f}$ and the mixed teacher vectors $\bm{t}$.
By changing the number of layers of the Siamese network and the third CNN, we can introduce the mixup in the middle layers.
As a similar method, Vikas Verma et al. \cite{manifold mixup} proposed "manifold mixup".
They proposed a method that randomly switch the place of the layer to perform mixup.

In the prediction phase, one of the sub-networks and the CNN are directly connected as one network and the class of the input image is estimated by feeding the input image to the input layer of the connected network.

We can extend the mixing of two feature maps to three or more.
For example, to extract three feature maps, we can used the triplet network architecture \cite{triplet} instead of the Siamese network.
The mixing weights can be generated using the Dirichlet distribution instead of the $\beta$ distribution.

\section{Experiment}

\subsection{Dataset and Network Architectures}


To confirm the effectiveness of the proposed mixup of feature maps in the hidden layer, we have performed experiments to compare the classification accuracies and the obtained feature maps in the hidden layer by using CIFAR10 dataset.
CIFAR10 dataset includes 60,000 labeled small images for image classification. 
The size of each image is $32 \times 32$ pixels. 
The number of classes is 10.
Usually they are divided into the training samples with 50,000 images (5,000 images per each class) and the test samples with 10,000 images (1,000 images per each class).
In the following experiments, the number of training samples was reduced to 5,000 (500 samples per each class) to make the improvement of the generalization ability of the mixup approaches clear.
The 10,000 test samples were used for evaluation of the classification accuracy.

In the proposed method, the feature maps from the first and the second convolution layers were extracted. 
To extract the feature maps from the first convolution layer, the Siamese network and the triplet network with only one convolution layer were used.
The networks with two convolution layers were used to extract the feature maps from the second convolution layer.

For the comparisons, we also evaluated the classification accuracies of the network trained without mixup and the network trained with the original mixup.

To train the parameters of the networks, the standard stochastic gradient descent (SGD) learning algorithm were used for all the network architectures.
The learning rate of SGD was started at 0.01 and was gradually reduced by multiplying 0.1 after 100 epochs.
In addition, ReLU function is used as activation function for output of each convolution layer and output of hidden layer of Fully-connected layer.
To prevent the over fitting to the training samples, we used the weight decay.
The weight decay parameter was changed depending on the over fitting.
For the original CNN without mixup the weight decay parameter was set to 0.04 and it was set to 0.02 for the mixup of two images or two feature maps.
In addition, it was set to 0.02 for the mixup of three feature maps of the second convolution layer.
The weight decay parameter 0.01 was used for the mixup of three images or three feature maps of the first convolution layer.
The mixing parameter $\alpha$  in the mixup was changed from $0.2$ to $1.0$.


\subsection{Comparison of the Classification Accuracy}

The classification accuracy of the model is shown in Table \ref{fig:acc}.
In this table, "cnn (original)" and "cnn (mixup)" denote the network that was trained without mixup and the network that was trained with the mixup of the two images, namely the original mixup.
The mixup with the three input images is also denoted as "cnn (mixup3)".
The proposed mixup of the two feature maps in the hidden layers are shown as "cnn (conv1-mixup)" and  "cnn (conv2-mixup)" where "conv1" and "conv2" are used for the feature maps extracted from the first convolution layer and the second convolution layer respectively.
So "cnn (conv1-mixup)" means that the network is trained by using the mixup of two feature maps extracted the first convolution layer.
Similarly the proposed mixup of the three feature maps are denoted as "cnn (conv1-mixup3)" or "cnn (conv2-mixup3)".

\begin{figure}[ht]
\begin{center}
\begin{tabular}{|c|c|c|c|} \hline
model & accuracy & $\alpha = 0.7$ & $\alpha = 1.0$ \\ \hline \hline
cnn (original) & $50.01\%$  & $\times$ & $\times$ \\ \hline
cnn (mixup) & $\times$ & $52.00\%$ & $52.22\%$ \\ \hline
cnn (mixup3) & $\times$ & $51.39\%$ &$52.22\%$ \\ \hline
cnn (conv1-mixup) & $\times$ & ${\bf 52.39}\%$ & ${\bf 52.93}\%$ \\ \hline
cnn (conv2-mixup) & $\times$ &$51.69\%$ & ${\bf 52.26}\%$ \\ \hline
cnn (conv1-mixup3) & $\times$ &${\bf 54.67}\%$ &${\bf 55.31}\%$ \\ \hline
cnn (conv2-mixup3) & $\times$ & ${\bf 52.78}\%$ & $51.50\%$  \\ \hline
\end{tabular}
\end{center}
\caption{accuracy of classification}
\label{fig:acc}
\end{figure}

From the comparison experiments, we observed that the effect of mixup increase for this dataset as the value of $\alpha$ becomes larger.
Namely the accuracies for the cases where the mixing parameter $\alpha$ was less than $0.7$ were less than the cases with $\alpha=0.7$ or $\alpha=1.0$.  
In the this Table \ref{fig:acc}, we show only the accuracies for the cases with $\alpha=0.7$ and $\alpha=1.0$.

From this Table, it is noticed that the classification accuracies of "cnn (conv1-mixup3)" are highest for both cases with $\alpha=0.7$ and $\alpha=1.0$.
Also, the classification accuracies of "cnn (conv1-mixup2)" are better than the original mixup of two images "cnn (mixup)".
These results shows that the mixup of the feature maps is more effective than the mixup of the input images.

For any $\alpha$, the accuracy of the model "cnn (conv2-mixup)" in which mixup is applied to the feature maps extracted from the second convolution layer is almost same as the "cnn (mixup)" or "cnn (mixup3)". 
In addition, the accuracy of the model "cnn (conv2-mixup3)" is lower than the "cnn (mixup)" or "cnn (mixup3)".

It is expected that the first convolution layer is working to suppress the general image variations and the second convolution layer is extracting more class specific information.
Thus we think that the mixup of the feature maps in the first convolution layer can generate the reasonable intermediate feature maps but the mixup of the feature maps in the second convolution layer maybe destroy the class specific information.

\subsection{Comparison of the Feature Maps in the Hidden Layer}

\begin{figure}[ht]
\begin{center}
\includegraphics[scale=1]{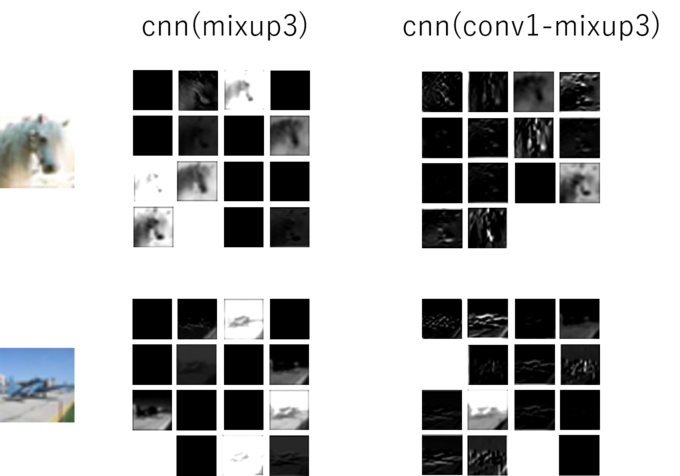}
\caption{feature map extracted by the first convolution layer}
\label{fig:featuremap}
\end{center}
\end{figure}

Furthermore, we compare the feature maps extracted by the first convolution layer of the model "cnn(mixup3)" and "cnn(conv1-mixup3)". 
Figure \ref{fig:featuremap} shows the visualization of the extracted feature maps for the same input images.

From this figure, we can find that the features map obtained by the "cnn (mixup3)" model are influenced by the brightness of the original image.
On the other hand, the shape edges are extracted in the feature maps obtained by the "cnn (conv1-mixup 3)" model regardless of the brightness of the original image.
This result shows the some improvements of the feature map in the first convolution layer by the mixup of the feature maps.

\section{Conclusion and Future Works}


In this paper, we proposed an data augmented learning algorithm in which the feature maps in the hidden layer are mixed.
To extract the feature maps during the training, the Siamese network or the triplet network architecture is used.
Experimental results show that some improvement of the classification accuracy is achieved by applying the mixup to the feature maps extracted from the first convolution layer. 
On the other hand, applying mixup to the second convolution layer does not produce significant improvement. 
From the experiment results, it is noticed that the classification accuracy depends on the mixing parameter $\alpha$ but the tendency is the almost same for all the mixup models for CIFAR10 dataset.

In this paper we did not consider the effect of the distances between the mixinig images or feature maps but we think that distances probably are important factor to generate good intermediate image or feature maps.
So we would like to introduce some mechanism to control the probability of the mixup depending on the distances.

\section*{Acknowledgment}

This work was partly supported by JSPS KAKENHI Grant Number 16K00239.

\end{document}